# Holographic Automata for Ambient Immersive A. I. via Reservoir Computing

## Theophanes E. Raptis[a,b,c]

[a]National Center for Science and Research "Demokritos", Division of Applied Technologies, Computational Applications Group, Athens, Greece; [b]University of Athens, Department of Chemistry, Laboratory of Physical Chemistry, Athens, Greece; [c]University of Peloponnese, Informatics and Telecommunications Dept., Tripolis, Greece
E-mail: rtheo@dat.demokritos.gr

**Abstract:** We prove the existence of a semilinear representation of Cellular Automata (CA) with the introduction of multiple convolution kernels. Examples of the technique are presented for rules akin to the "edge-of-chaos" including the Turing universal rule 110 for further utilization in the area of reservoir computing. We also examine the significance of their dual representation on a frequency or wavelength domain as a superposition of plane waves for distributed computing applications including a new proposal for a "Hologrid" that could be realized with present Wi-Fi/Li-Fi technologies.

## Introduction

Distributed computing has a long history running the full half of the previous century and its development went hand in hand with the rise of connectionist paradigm out of the study of both natural and artificial neural networks [1]. Perhaps one of the first application of holographic principles in general computation are to be found in the entirely original method of tearing by Kron [2], later termed "Diakoptics", which was invented for the efficient solution of large electrical networks by decomposition. This was initially formalized by Roth [3], [4], [5] and later analyzed by Weinzweig [6] and others [7], [8], [9] in topological terms based on a homology-cohomology chain pair revealing the existence of so called, "twisted isomorphisms" between kernels and cokernels at successive chain levels. Its intrinsic relationship with a version of Huygen's principle and holography were further elaborated in the work of Bowden [10], [11], [12]. Similar thoughts can be traced at the foundations of General Systems Logical Theory (GSLT) by Jessel and Resconi [13], [14], [15] while another direction appears in the analysis of distributed computation as a hierarchy by Manthey [16],[17],[18].

At about the same time these original works appeared, parallels at the corssroads of the emerging cognitive science and the connectionists school gave additional fruits as in Smolenski [19], [20] who introduced Kronecker products in an effort to merge symbolic and conceptual representations in neural networks, Pale [21], [22], [23] who introduced holographic distributed representations of compositional data structures based on a type of convolution algebra. An interesting parallel is that of Kanerva's Binary Spatter Codes (BSC) [24][25][26]. More recently, Galant and Okaywe [27] used vectorized knowledge representations. There is also a stated equivalence of the previous models with Geometric Algebra due to Aerts [28]. Aerts and Gabora [29], [30] were also able to present an equivalent generic theory of concepts on Hilbert spaces based on the SCOP model, a generalization of quantum mechanics formalism [31],[32].

The previous approaches, although independently developed, they can also be studied in comparison with the recent adoption by Jaeger [33], [34], [35] of the "Conceptor" principle, a kind of abstract idea (concept) formed by a network in an effort to deepen understanding of the power of the so called, reservoir computing model [36], [37], [38]. In this new paradigm exceeding the power of standard neural networks, great flexibility is offered by an increase of dimensionality due to the introduction of a reservoir holding a great number of instants from the trajectory of an appropriate high dimensional dynamical system which could also be a recurrent neural network. The trajectory is also influenced by external perturbation coming for simple, linear networks accepting inputs from sensors while a second linear network performs a sampling of the resulting evolution to extract the outputs via a simple regression scheme. Introduction of the conceptor principle may also allow for a new, dynamic type of knowledge representations since a kind of logical calculus is enabled between them. This is strongly associated to the recently becoming acute problem of interpretability, due to the lack of complete understanding of how modern, extremely complex AI entities reach their conclusions [39]. It would be of great interest to know if there are measures able to discern the formation of a conceptron configuration in a holographic network of distributed computations.

Recently, Yilmaz [40], [41], [42] introduced an interesting simplification with the realization of cellular automata reservoir computing (ReCA). Bye [43] as well as Nichele and Molund [44], [45] continued along similar lines verifying the power of the ReCA model for deep learning. In these works, the reservoir is represented by many instances of one or more 1D or 2D CA of which the rules have been chosen close to the so called, "edge-of-chaos" area which are directly associated with universal computing power like the already proven Turing universal, rule 110 [46] and the "Game-of-Life" [47]. In the next section we apply a transformation of the original proposal by Yilmaz, by developing a general procedure for the direct decomposition of any 1D CA configuration into a set of orthogonal indicators for each specific neighborhood which is of theoretical interest rather than practical computations while in section 3, we develop a more economical approach towards a semi-linearization of the dynamics with the previously found projectors serving to correct the remaining nonconforming parts. Our interest is in extracting afterwards a particular translation of the dynamics in the frequency or wavelength domain for applications in distributed and reservoir computing according to the previously exposed paradigm. In section 4, we provide an alternative proposal for distributed computations based on a multi-agent system of active routers which serve as the reservoir in the sense originally proposed by Yilmaz. We close by further examining the significance of the feasibility of such ambient systems.

**Neighborhood specific orthogonal decomposition**

In this and subsequent section we shall make use of the following notations. An all zeros and all ones vectors shall be written as $\mathbf{0}$ and $\mathbf{1}$ while $I$ shall denote the identity linear operator. The sum of digit of a binary vector shall be denoted as $\Sigma(\mathbf{x})$. We shall also denote the complement of a binary vector $x$ as $\overline{\mathbf{x}} = \mathbf{1} - \mathbf{x}$, the translation from a $\{0, 1\}$ to a Hadamard encoding in $\{-1, +1\}$ via $\mathbf{x}^* = 2\mathbf{x} - \mathbf{1} = \mathbf{x} - \overline{\mathbf{x}}$ and the application of a DFT Vandermond matrix with $\tilde{\mathbf{x}} = \mathbf{F} \cdot \mathbf{x}$. A rank 1 matrix for a circulant convolution filter $\boldsymbol{C}$ acting on a finite signal with infinite periodic repetition shall be denoted with the aid of a convolution mask as ( $c_1,\ldots,c_k$ ) while filter construction over any finite vector of length $L$ involves a single defining vector $\mathbf{c} = \begin{bmatrix} c_1, c_2, \ldots, c_k, 0, \ldots, 0 \end{bmatrix}$ with a left cyclic shift. In Matlab/Octave notation, the total operation is given via a degenerate Toeplitz matrix using the command *toeplitz*($c$(1), fliplr( $c$(2:end) ), c).

In previous work [46] we proposed a particular decomposition for all CA dynamics in a pair of linear and non-linear maps the first being realized with simple circulant filters. To complete the representation and contrast it to the one presented later on, we show how to construct a complete, analytical representation out of these two maps. We restrict attention to 1D CA defined by a lattice of binary valued $L$ cells and a transition rule which can always be compressed into a single integer as $(r)_2 = (\sigma_0,...,\sigma_{2^{|N|}-1})$ where $|N| = k_L+1+k_R$ for an interaction neighborhood of radii $k_L$ and $k_R$ for a number of left and right bits in general. Standard, elementary CA correspond to the choice $k_L = k_R = 1$ but our treatment in this section shall be valid for all cases. Any possible neighborhood configuration belongs to the dictionary of all strings or Hamming space [47] which under LEX ordering correspond 1-1 to the integers in the interval $\left[0,...2^{|N|}-1\right]$ via the polynomial representation uniquely marking each pattern.

The total operation for any 1D binary array $s$ can be given with the aid of a special polynomial map $p_R$ and a circulant matrix defined by a vector $\mathbf{c} = (1,2,...,2^{|N|},0,...,0)$ for an arbitrary neighborhood $|N|$ as

$$\mathbf{x}_{n+1} = p_R\left(\mathbf{1} + \mathbf{C} \cdot \mathbf{x}_n\right) \qquad (1)$$

The polynomial map is to be defined as follows. Given that the interior filter $h = \mathbf{C} \cdot \mathbf{x}$ is only responsible for creating an "addressing" or "pointer" space for the transition rule in the codomain $[0,\ldots,2^{|N|}-1]$. Any transition can be realized either as a zero or a shifted zero of an arbitrary polynomial which then adopts the natural representation

$$p_R(z) = \prod_{i=1}^{S_2(|R|)}\left(z - h_i^0\right) \prod_{j=1}^{|N|-S_2(|R|)}\left((z-1) - h_j^1\right) = \sum_{m=0}^{|N|-1} a_m z^m \qquad (2)$$

In (2), we identify $S_2(|R|)$ with the sum of bits in the binary representation of any transition rule $|R|$ and $\{h_i^0, h_j^0\}$ with the addresses of each separate zero or one in this expansion respective so as to collectively pass all activations from the inner filter during any iteration. Notably, due to the fundamental property of circulant filters [48] one also has the diagonalizing property of DFT matrices as $\mathbf{C} = \mathbf{F}^{-1} \cdot \Lambda \cdot \mathbf{F}$ with $diag(\mathbf{C}) = \widetilde{\mathbf{c}}$ which allows writing a dual, distributed representation in the frequency or wavelength domain as

$$\widetilde{h}_{n+1} = \Lambda \cdot \mathbf{F}\left(p_R(\widetilde{h}_n)\right) = a_0\delta(\omega) + \sum_{m=0}^{|N|-1} a_m (i\omega)^m \Lambda \cdot \left(D_\omega^m\right)\widetilde{h}_n \qquad (3)$$

In what follows, we shall attempt a different approach for 1D CA, in an effort to approximate as much as possible a unitary approach with a minimal perturbation, especially for the class of certain Turing universal rules as rule 110. According to recent work by Riedel and Zenil [49] it is possible to find methods by which universality can be emulated via the introduction of certain special patterns also termed, "compilers" that serve to translate the behavior of a 1D CA with another. This severely broadens the notion of universality which becomes then ubiquitous given appropriate pattern classifiers that could also be stated in terms of certain intelligent output filters as those provided by advanced regression filters met in the general Machine Learning (ML) and Neural Network literature.

We next consider a method by which a complete, pattern specific set of projection operators $\hat{P} : \mathbf{B}^L \to \mathbf{B}^L$ where $\boldsymbol{B}$ stands for the unit hypercube, can be constructed such that each neighborhood pattern in a sequence of L bits is uniquely mapped into a new Boolean vector so as to obtain a new set of $n_N \leq 2^{|N|}$ indicator vectors.

Let then $\boldsymbol{C}$ stand for an arbitrary circulant filter of which the $L$ x $L$ matrix representation is always decomposable to exactly $L$ permutations as $C = \sum_{m=1}^{L} c_m \Pi_C^m$ where $\Pi_C$ the elementary, cyclic permutation matrix. The action of each elementary filter $K_m = \Pi_C^m$ is to recognize the presence of a $x_m = 1$ bit in position $m$ of any Boolean vector $\boldsymbol{x}$. Obviously, any zero position can also be recognized via the application of the same filter over the complementary vector $\overline{\mathbf{x}}$. This immediately leads to a generalization where any arbitrary pattern $x_{i-k} x_{i-k+1} \ldots x_{i+k}$ can be effectively recognized via a direct bitwise AND product (*) of the resulting vectors from each filter which is the equivalent of componentwise multiplication for Boolean variables with $K \cdot \mathbf{1} = \mathbf{1}$, $\mathbf{x} * \mathbf{1} = \mathbf{1} * \mathbf{x} = \mathbf{x}$. This then suggest the following definition.

$$\hat{P}_{x_1 \ldots x_N}(\mathbf{x}) = \left( K_{-k_L} \cdot \mathbf{x} \right) * \ldots * \left( K_{-k_R} \cdot \mathbf{x} \right) \tag{4}$$

In (1), the succession of filters is defined via the corresponding masks following a sequence of bit shifts from the lower left position to the highest right one, while the central cell position is always assigned the identity operator $I$. In the simplest case of a symmetric neighborhood of elementary CA we simply have the triple $\left\{ K_-, I, K_+ \right\}$ with the corresponding masks $\{(1, 0, 0), (0, 1, 0), (0, 0, 1)\}$. For example, the pattern 010 would have the equivalent projector

$$\hat{P}_{010}(\mathbf{x}) = \left( K_- \cdot \overline{\mathbf{x}} \right) * \left( I \cdot \mathbf{x} \right) * \left( K_+ \cdot \overline{\mathbf{x}} \right) \tag{5}$$

We notice that application over complements leads to a multinomial expansion since $K_m \cdot \overline{\mathbf{x}} = K_m \cdot \mathbf{1} - K_m \cdot \mathbf{x} = \mathbf{1} - K_m \cdot \mathbf{x}$ thus (5) can be expanded as

$$\hat{P}_{010}(\mathbf{x}) = \left( \mathbf{1} - K_- \cdot \mathbf{x} \right) * \left( \mathbf{1} - K_+ \cdot \mathbf{x} \right) * \mathbf{x} = \mathbf{x} - \left( K_\pm \cdot \mathbf{x} \right) * \mathbf{x} + \left( K_- \cdot \mathbf{x} \right) * \left( K_+ \cdot \mathbf{x} \right) \tag{6}$$

where $K_\pm = K_- + K_+$. As a final result, the set of all projectors must satisfy a kind of resolution of the identity of the form

$$\sum_{m=0}^{2^{|N|}-1} P_{(m)_2}(\mathbf{x}) = \mathbf{1} \tag{7}$$

The resulting set of all such $2^{|N|}$ possible indicator vectors are naturally orthogonal with respect to the bitwise AND product and its numerical values via the polynomial representation can also be found for any global map over all possible $\boldsymbol{x}$ states in $[0, \ldots, 2^L]$ to correspond to a particular fractal matrix. From the theory of digital adders formed by a recursive combination of AND and XOR operators we know that zeros of any bitwise AND matrices correspond to the case of XOR being an exact adder as $|\mathbf{x}| \oplus |\mathbf{y}| = |\mathbf{x}| + |\mathbf{y}|$ where $|\mathbf{x}|$ denotes here the exact numerical value corresponding to the particular binary pattern. This matrix is strictly associated with a deformation of the Hadamard matrix and has a recursive form via the so called, Langlet generator $h_0$ as

$$h_{n+1} \leftarrow \begin{bmatrix} h_n & h_n \\ h_n & \mathbf{0} \end{bmatrix}, \ h_0 = \begin{pmatrix} 1 & 1 \\ 1 & 0 \end{pmatrix} \tag{8}$$

The resulting matrix is shown in fig. 1, and it has been previously introduced in the Langlet transform that has been reviewed in a book by Zaus [50] on hypercubical calculus. All orthogonal components resulting from any decomposition by the projectors in (1) will necessarily have integer encodings belonging to pairs with positive values of (7).

Next, we introduce a total evolution operator $\hat{U} : \mathbf{B}^L \rightarrow \mathbf{B}^L$ in the state space of any 1D CA as

$$\hat{U} = \sum_{m=1}^{2^{|N|}-1} \sigma_m^r \hat{P} \tag{9}$$

With the aid of (9) any 1D CA evolution can now be written in the form

$$\mathbf{x}_{n+1} = \hat{U}(\mathbf{x}_n) = \sum_{m=1}^{2^{|N|}-1} \sigma_m^r \hat{P}(\mathbf{x}_n) \tag{10}$$

Next, assume we want to know the dual Fourier representation of (9) to a wavelength or frequency domain. We first notice that (10) contains essentially $2^{|N|}$ componentwise products of exactly $|N|$ elements and in analogy with the convolution theorem for function products any such product $y_1 \ldots y_N$ can be inverted with a set of convolution integrals over their Fourier transforms. At this point we can again make use the fundamental property for the diagonalization of circulant filters. The same logic when applied in (10) results in the total expression on a wavelength domain as

$$\tilde{\mathbf{x}}_{n+1}(k) = \sum_{m=1}^{2^{|N|}-1} (-)^{|N|-\Sigma_m} \sigma_m^r \left\{ \left( \lambda_{i-k_L} \cdot \tilde{\mathbf{x}}_n(\omega) \right) \otimes \ldots \otimes \left( \lambda_{i+k_R} \cdot \tilde{\mathbf{x}}_n(\omega - \omega^{(|N|)}) \right) \right\} \tag{11}$$

Some care is required regarding the sign factor $|N| - \Sigma_m$ where $\Sigma_m$ stands for the sum of digits of each neighborhood due to the need for inclusion of complements in the summand of (11). For each zero of a neighborhood we should have $F \cdot \bar{\mathbf{x}} = \tilde{\mathbf{1}} - \tilde{\mathbf{x}}$ where $\tilde{\mathbf{1}} = (L, 0, \ldots, 0)$ thus essentially shifting only the DC component (again sum of digits) of the original configuration apart from a sign change. With a minimal change of convention such that $x_n(1) = \sum x_n(i) - L$ the sign factor can be extracted outside the resulting multi-convolution integral. We note in passing that the expression reached has some analogies with the analysis of non-linear filters with multiple responses originally made by Wiener and Volterra [51], [52]. While the complexity of (11) appears to be exponential, care is taken in the next section to reduce this in only a single strictly linear contribution and a few correcting terms using only a minimal number of the total projectors possible.

Last but not least, it is possible to expand the projectors into the whole $b^{|N|}$ domain of any higher alphabet in radix $b$ for a 1D CA by promoting all symbols $\sigma_m \in [0,...b^{|N|}-1]$ in the associated set of $b$ roots of unity $\omega_m = \omega_0^m$, $\omega_0 = \exp(2\pi i/b)$ which is a straightforward generalization of the Hadamard code for the binary alphabet and taking convolution masks as $(0,...,\omega_m^{-1},...,0)$ to redefine each filter in the projectors as $\hat{P}: \mathbf{C}^L \to \mathbf{B}^L$ via the integer part of each $\text{Re}(K_m \cdot \mathbf{x})$. In the next section we examine an approach that leads to a partial linearization of the dynamics where a minimal number of projectors are used for the non-linearities using examples associated with Turing universality and computation at the edge of chaos.

**Split linearization methods for "Edge of Chaos" computation**

Ideally, we would like to find a convolution mask $(c_1, c_2,...,c_{|N|}, 0...)$ such as to have a direct translation of the dynamics into the form $\mathbf{x}_{n+1} = C \cdot \mathbf{x}_n$. Such a feat is not possible for the cases of interest as for instance, the Turing universal rule $(110)_2 = (0, 1, 1, 1, 0, 1, 1, 0)$ since universality is related to a type of quasi-chaotic, generally undecidable behavior that would require at least exponential resources to fully linearize as for instance, in the case of endless memory increase. Yet, it is possible to find types of multi-dimensional indexing that allow splitting the original into a mostly linear part and a minimal remnant that can be effectively treated with the projectors defined in the previous section.

Before entering the specifics of the method we will make an important distinction between two sets of rules of which one can be truncated into a ternary alphabet for easing the desired split. This second set is associated with the well known Langton $\lambda$ parameter [53] or simply the sum of digits of each rule via $\lambda = \Sigma(r)/2^{|N|}$. In particular, any rule may have an associated set of $k$ transitions as well as $2^{|N|} - k$ idle states where cell values do not change. We can then posit a new, truncated rule form where a 1 is added to a 0 cell state or subtracted by a previous 1 position with all other positions set to 0 which naturally leads to a ternary alphabet. It is possible to automatically perform the truncation via a bitwise logical formula given that the middle bits on any LEX arrangement of strings for a 3 cell neighborhood correspond to the integer $(204)_2 = (0, 0, 1, 1, 0, 0, 1, 1)$. We than have

$$(r_T)_2 = (r \oplus 204)_2 - 2(\bar{r} * 204)_2 \tag{12}$$

The bitwise XOR extracts the non-idle cell values as differences while the sign is taken care by a bitwise AND with the complement of $r \to 255 - r$ which simply inverts positive ones when they correspond to a 1 $\to$ 0 transition. Computing a set of truncated rules over the whole set $[0,...,255]$ allows finding the equivalent density of absolute values from $|r^T|$ for $\lambda_T$ and characterize the efficiency of this scheme via the ratio $\lambda/\lambda_T$. The result is shown in figure 2(a-b) where only a small subset of 80 rules (31.25%) seems to be appropriate for this technique being efficient in an economy of transitions. In the case of rule 110, we have the equivalent $(0, 1, 0, 0, 0, 1, 0, -1)$ for which the dynamics can be written as

$$\mathbf{x}_{n+1} = \mathbf{x}_n + r^T(\mathbf{x}_n) \tag{13}$$

We notice that the original 8$^{th}$ order polynomial map defined by (2) for the 110 rule is here defined by the set of coefficients [134400, -319360, 291716, -140289, 39729, -6870, 714, -41, 1] while an analogous construction would be possible for $r^T$ in (10) with both +1 and -1 shifts of zeros.

To proceed further we need a suboptimal solution to the linear optimization problem of finding a convolution mask absorbing the maximal possible set of transitions. This is equivalent to the below optimization problem where $\sigma_1\sigma_2\sigma_3$ are Boolean variables running over all possible neighborhood configuration patterns and $c_i$ are the mask's integer variables in an arbitrary range [-N, N]$^3$ as

$$\arg\min\left\{\left\|\sum_{i=1}^{3}c_i\sigma_i^{j}-r_j^{T}\right\|\right\}_{j=0}^{j=2^{|N|}-1} \tag{14}$$

That is we ask for a triplet of mask values capable of projecting each and every one of the unique combinations of three Booleans into the appropriate variables of the truncated rule. Such a problem cannot be globally solvable with only three unknown integer variables. We also mention that promoting the mask values into rationals like $c_i = p_i/q_i$ only results after simplification into a set of four integer parameters with $p_i$ and $q_i$ as factors which is still not globally solvable. After trial and error one finds that one of the closest possible approximations is given by a mask $c_T = (1, 0, 0)$ which only gives the (*mod 2*) sequence $r_{mod2}$ for all neighborhoods and of which the distance is $|r^T - r_{mod2}| = 2$ at positions 3 and 7 hence, one can rewrite (13) as

$$\mathbf{x}_{n+1} = \left(I + C^T\right)\cdot\mathbf{x}_n + \varepsilon_T(\mathbf{x}_n) = K\cdot\mathbf{x}_n + \varepsilon_T(\mathbf{x}_n) \tag{15}$$

The new convolution mask for $K$ is simply (1, 1, 0) while the correction vector $\varepsilon$ can be found in terms of the projectors of section *2*. From the exact positions of deviation we then obtain a linear combination $\varepsilon \sim \alpha\hat{P}_{110} + \beta\hat{P}_{111}$. Comparing against the exact form of $r^T$ then immediately results in $\alpha = -1, \beta = -2$ so that we finally obtain

$$\mathbf{x}_{n+1} = K\cdot\mathbf{x}_n - \hat{P}_{110}(\mathbf{x}_n) - 2\hat{P}_{111}(\mathbf{x}_n) \tag{16}$$

The Fourier dual of (16) can now be immediately derived following the general recipe given at the end of section *2* as

$$\tilde{\mathbf{x}}_{n+1}(k) = \Lambda^K\cdot\tilde{\mathbf{x}}_n(k) + \tilde{\mathbf{x}}_n(k)\otimes\left(\lambda_1\cdot\tilde{\mathbf{x}}_n(\omega-\omega')\right)\otimes\left(\lambda_2\cdot\tilde{\mathbf{x}}_n^{NOT}(\omega-\omega'')\right) - ...$$
$$2\tilde{\mathbf{x}}_n(k)\otimes\left(\lambda_1\tilde{\mathbf{x}}_n(\omega-\omega')\right)\otimes\left(\lambda_2\tilde{\mathbf{x}}_n(\omega-\omega'')\right) \tag{17}$$

In (17), $\lambda_1$ and $\lambda_2$ stand for the diagonal forms of the $K_+$ and $K_-$ filter matrices in the case of a symmetric 3-neighborhood while the sign inversion in the inverted $\hat{P}_{110}$ term follows the convention of section 2 for the necessary change of the DC component. The inverted form can be seen as a single composite signal passing through a pair of linear filter cascades with appropriate bandwidth which could even be made to correspond to some appropriately layered material. More on possible realizations follow in the next section.

Other examples of potentially useful rules for "edge of chaos" computing in the reservoir model can be found via the study of Lyapunov exponents [54], [55]. One such rule with interesting behavior as reported in [55] is the rule $(106)_2 = (0, 1, 0, 1, 0, 1, 1, 0)$. This part from the previous example in that it does not belong to the subset of efficient truncations thus it has to be used as is for the optimization scenario in (14). We immediately observe from its binary expansion that it satisfies $|r^T - r_{mod2}| = 2$ for positions 6 and 7 hence it can be translated exactly as in (16) this time with the choice $c = (1, 0, 0)$, $\alpha = 1$, $\beta = -1$ and a change of the first projector into $P_{011}$ so that its explicit form becomes

$$\mathbf{x}_{n+1} = C \cdot \mathbf{x}_n + \hat{P}_{011}(\mathbf{x}_n) - \hat{P}_{111}(\mathbf{x}_n) \tag{18}$$

Translation on the frequency/wavelength domain follows as before. In the last section we examine possible applications of the reformulation of the dynamics in the field of reservoir computing along the lines already described in the introduction.

**Active Router Multi-Agent Systems (ARMAS) and the Hologrid proposal**

In an ordinary understanding of distributed computation, one would expect each separate mode in expansions like that of (17) being separately realized by independent parts of an entire system with sufficient number of complementary units responsible for updating the total spectrum via delayed measurements with a depth according to the available bandwidth. This poses a problem with respect to memory requirements, yet there is a much simpler implementation utilizing the fact that such frequency expansions allow a reinterpretation of any CA as a single signal compressed representation of the state of a whole array. There are several extensions to this scheme for spectral based, analog computing on which we have commented in a previous work [56]. Methods for effectively encoding the coefficients of a spectral representation are readily available in telecommunications, mobile telephony and optical fibers as in the case of Wavelength Division Multiplexing (WDM) [57] or Orthogonal Frequency Division Multiplexing (OFDM) [58] which bring about the additional ingredient of a certain similarity with Hilbert spaces with the inner product being replaced by a temporal average.

Regarding the reservoir computing proposal, it is in principle possible to create many dispersed local reservoirs using a multi-agent approach [59]. We shall then have to introduce a type of "active router" which does not merely retransmit messages via some address lookup table. Instead, any such is equipped with a bank of all-pass, ideally lossless filters [60], [61] of which a cascade realizes a complete update step for the incoming signal. Assuming also an appropriate tagging system indicating the next agent to act as a receiver, the totality of such agents realizes a complete cycle keeping a set of trajectory instances as $\{x_n, x_{n+1}, \ldots, x_{n+m}\}$ where $m$ the number of local agents available. It is for long known in algorithmic theory that many other computing operations can always be translated into single loop, branchless instructions [62] although the particular articulations are beyond our scope.

Such a local collection of agents can always be supported by a set of nearby sensors and simple one layer regression modules to feed appropriate perturbations into some of the instances of any trajectory so as to realize a complete reservoir computing unit capable of locally analyzing the flow of materials, vehicles or other human activities. A perhaps self-similar hierarchy of such units could then be made to extend across a larger environment as a kind of symbiotic, cybernetic organism, cross-feeding between different layers of abstractions. A complete study and simulation of a similar system will require a separate series of articles. Lastly, we mention one additional theoretical possibility that has been touched at a simpler level in a previous work [63]. In principle, one can theorize on any sufficiently noisy background as being infinite dimensional. It then presents us with the possibility of using it as a natural reservoir given appropriate filtering techniques and table lookups for isolating interesting substructures to be used as the set of some trajectory instances. Notably, the arithmetized version of a multitude of Turing machines (TM) possible as given in [63], allows a direct transfer into an OFDM/OWDM protocol such that an ideal, infinite bandwidth signal or equivalently, any TM/UTM working on a (*mod p*) with cyclic boundary conditions can be realized in the Fourier domain using simply additive waveform synthesis techniques to replace any read/write operations. Moreover, Gaussian noise spectra can be classified as the infinite limits of successive convolutions [64] which could allow for a specialized application of certain holographic representations directly upon the noisy substrate offered by the broadband EM spectrum. An appropriate use of convolution and correlation algebras [65] may then allow for the imposition of certain pseudo-Hamiltonian structures the details of which will be presented in subsequent work.

**Discussion and conclusions**

We report a new technique for the neighborhood decomposition of any elementary 1D CA which can be utilized for the replacement of the analytic expression of any such CA dynamics of polynomial complexity. The resulting tool is then used to simplify the dynamics in a dual Fourier domain, standing for a holographically implemented CA via a split linearization method which leads to a special class of multi-convolution kernels. We comment on the combined use of this technique with a recent proposal of a CA based reservoir computing for computations near the edge-of-chaos capable of supporting deep learning operations and we delve into the technical details necessary for the realization of similar distributed computing as an ambient, immersive A.I. using appropriately defined "active routers" and filter banks. Possible beneficial uses include industrial applications and self-organized factories as in future space colonies or asteroid mining operations, underwater and deep sea operations with appropriate modifications for the acoustic spectrum, and similar.

We should stress the fact that realization of similar systems may also lead to enhanced surveyance mechanisms with the capacity of real time decision making, drone swarm cooperation and guidance and other similar possibilities. As these may contravene recent protests and forewarnings on the future of privacy, human rights and other sensitive matters, the author would like to ensure that the present report neither addresses nor endorses any such use for extreme policing or similar and only the theoretical and technical feasibility is here reported. On the other hand, the present state of evolution and the overall computing power increase in the relevant fields call for deeper examination of the complexities and the ramifications involved in the massive interaction of humans and distributed machine intelligence, as already requested by the so called, "friendly A. I." [66], movement.

**References**

[1] McLelland, J. L., Rumellhardt, D. E., (1986) *Parallel Distributed Processing*, Volume II, Cambridge, Mass: MIT Press.

[2] Kron G. (1963) *Diakoptics: The Piecewise Solution of Large Scale Systems*, MacDonald Publishing.

[3] Roth, J.P. (1959) "An application of algebraic topology to numerical analysis: On the existence of a solution to the network problem", *PNAS* 41(7):518–21

[4] Roth, J.P. (1959) "The validity of Kron's method of tearing", *PNAS* 41(8):599–600

[5] Roth (1959) "An application of algebraic topology: Kron's method of tearing", *Quarterly of Applied Mathematics* 17:1–24

[6] Weinzweig, A. I., (1960) "THE KRON METHOD OF TEARING AND THE DUAL METHOD OF IDENTIFICATION", *Quart. Appl. Math.* 18(2):183-190.

[7] Harrison, B. K., (1963) "A discussion of some mathematical techniques used in Kron's method of tearing" *J. Soc. Industr. Appl. Math.* 11(2):258-281.

[8] Sehmi, N. S. (1986) "The Lanczos algorithm applied to Kron's method" *Int. J. Num. Meth. Eng.* 23:1857-1872.

[9] Maurice, O., Reineix A. *et al.*, (2014) "Kron's method and cell complexes for magnetomotive and electromotive forces" *Int. J. Appl. Math.* 44(4):183-191.

[10] Bowden, K, (1990) "On general physical system theories" *Int. J. Gen. Sys.* 18:61-79.

[11] Bowden, K, (1994) "Hierarchical tearing: an efficient holographic algorithm for system decomposition" *Int. J. Gen. Sys.* 23):23-37.

[12] Bowden, K, (1998) "Huygen's principle, physics and computers"*Int. J. Gen. Sys.* 18(1-3):9-32.

[13] Resconi, G., Jessel, M. (1986) "A general logical systems thoery" *Int. J. gen. Sys.* 12:159-182.

[14] Resconi, G., Auger, P. (1990) "Hilbert space and dynamical hierarchical systems" *Int. J. Gen. Sys.* 16:235-252.

[15] Resconi, G., Rattray, C. *et al*. (1999) "The language of general systems logical theory" *Int. J. Gen. Sys.* 28(4-5):383-416.

[16] Manthey, M. (1997) "A combinatorial bit bang leading to quaternions" *Proc. Helsinki Conf. Emergence, Complexity, Hierarchy, Organization,* ECHOIII.

[17] Manthey, M. (1998) "Distributed computation as hierarchy" *Comp. Sci. Dept. Tech. Report R-98-5005*, Aalborg Univ., preprint at arXiv:cs/9809019 [cs.DC]

[18] Manthey, M. (1998) "Distributed computation, the twisted isomorphism and auto-poiesis" *1st Int. Conf. Comp. Anticip. Sys.*, CASYS '97, Ed. D. Dubois, Liege, Belgium.

[19] Smolenski, P. (1990) "Tensor product variable binding and the representation of symbolic structures in connectionist systems" *A. I.* 46:159-216.

[20] Smolenski, P. (2012) "" *Phil. Trans. R. Soc. A* 370:3543-3569.

[21] Plate, T., (1991) "Holographic reduced representations: convolution algebra for compositional distributed representations" *Proc. 12th Int. J. Conf. A. I.* Ed. J. Mylopoulos, R, Reiter, M. Kaufmann, San Mateo, CA.

[22] Plate, T., (1995) "Holographic reduced representations" *IEEE, Trans. Neural Net.* 6(3):623-641.

[23] Plate, T. (2003) "Holographic reduced representations: Distributed representation for cognitive structures" Lecture Notes, Center for the study of language and information, Leland Stanford J. Univ., CSLI publications.

[24] Kanerva, P., (1988) "*Sparse Distributed Memory*" MIT Press.

[25] Kanerva, P. (1996) "Binary spatter-coding of ordered *K*-tuples ", *Art. Neural Net. Conf. ICANN96,* pp:869-873.

[26] Kanerva, P. (1998) "Large patterns make great symbols: an example of learning from example" in *Hybrid Neural Systems* pp:194-203.

[27] Gallant, S. I., Okaywe, T. W., (2013) "Representing objects, relations and sequences" *Neural Comp.* 25:2038-2078.

[28] Aerts, D., Czahor, M. *et al.*, (2006) "On geometric algebra representations of binary spatter codes" *Computer Research Repository*:arXiv:cs/0610075[cs.AI]

[29] Aerts, D., Gabora, L., (2005) "A theory of concepts and their combinations I-II" *Kybernetes* 34:167-221.

[30] Hampton, J. A. (2013) "Conceptual Combination: Extension and Intension. Commentary on Aerts, Gabora, and Sozzo" *TopiCS Comp. Sci.* 6(1):35-57.

[31] Aerts. D., (1983) "Classical theories and non classical theories as a special case of a more general theory" *J. Math. Phys.* 24:2441-2453.

[32] Busemeyer, J. R., Bruza, P. D., (2012) "*Quantum Models of Cognition and Decision*" UK:Cambridge Univ. Press.

[33] Jaeger, H. (2014) "Conceptors: an easy introduction" preprint in arXiv:1406.2671[cs.NE].

[34] Jaeger, H. (2014) "Controlling Recurrent Neural Networks by Conceptors" *Jacobs Univ. Tech. Report Nr 31.*

[35] Dima, A., (2015) "Generating animal or human locomotion patterns at different speeds, using conceptor morphing" thesis, Jacobs Univ. Bremen.

[36] Wolfgang, M., Nachtschlaeger, T. *et al.* (2002) "Real-time computing without stable states: A new framework for neural computation based on perturbations." *Neural Computation* 14(11): 2531–2560.

[37] (2007) "An overview of reservoir computing: theory, applications, and implementations" *Proc. Europ. Symp. Art. Neural Net. ESANN 2007*, pp. 471-482.

[38] Gallichio, C., Alessio, M. *et al.*, (2017) "Deep reservoir computing: a critical experimental analysis" *Neurocomputing* 268:87-99.

[39] Voosen, P. (2017) "The AI Detectives" *Science* 357(6346):22-27.

[40] Yilmaz, O. (2015) "Connectionist-Symbolic Machine Intelligence using Cellular Automata based Reservoir-Hyperdimensional Computing" preprint in arXiv:1503.00851 [cs.ET]

[41] Yilmaz, O. (2015) "Analogy Making and Logical Inference on Images using Cellular Automata based Hyperdimensional Computing" *Proc. NIPS Cogn. Comp.: Integrating Neural and Symbolic Approaches CoCo 2015.*

[42] Margem, M., Yilmaz, O. (2016) "An experimental study on cellular automata reservoir in pathological sequence learning tasks" *Proc. Art. Gen. Int.: $9^{th}$ Int. Conf. AGI 2016,* Springer.

[43] Bye, E. T., (2016) "Investigation of Elementary Cellular Automata for Reservoir Computing", MsC Thesis, Norw. Univ. Sci. Tech.

[44] Nichele, S., Molund, A. (2017) "Reservoir Computing Using Non-Uniform Binary Cellular Automata", *Complex Sys.* 26(3), preprint in arXiv:1702.03812 [cs.ET].

[45] Nichele, S., Molund, A. (2017) "Deep Reservoir Computing Using Cellular Automata" *Complex Sys.* 26(4), preprint in arXiv:1703.02806 [cs.NE].

[46] Raptis, T. E. (2016) "Spectral representations and global maps of cellular automata dynamics" *Chaos, Sol. Fract.* 91:503-510.

[47] Cohen, G., Honkala, I., Litsyn, S., Lobstein, A. (1997) "*Covering Codes*" Elsevier Sci.

[48] Davis, P. J. (1970) "*Circulant Matrices*" NY:Wiley.

[49] Riedel, J., Zenil, H. (2018) "Cross-boundary Behavioural Reprogrammability Reveals Evidence of Pervasive Universality" preprint in arXiv:1510.01671 [cs.FL].

[50] Zaus, M., (1999) "*Crisp and Soft Computing with Hypercubical Calculus*" Berlin:Springer-Verlag.

[51] Franz, M. O., Schölkopf, B. (2006) "A Unifying View of Wiener and Volterra Theory and Polynomial Kernel Regression" *Neural Computation* 18(12):3097-3118.

[52] Giannakis, G. B., & Serpedin, E. (2001) "A bibliography on nonlinear system identification" *Signal Processing* 81:533 – 580

[53] Langton, C. G. (1992) "*Life at the Edge of Chaos*" in *Artificial Life II*, Wesley.

[54] Baetens, J. M., De Baets, B. (2010) "Phenomenological study of irregular cellular automata based on Lyapunov exponents and Jacobians" *Chaos* 20:033112(1-15).

[55] Baetens, J. M., Gravner, J. (2014) "Introducing Lyapunov profiles of cellular automata" *Proc. $20^{th}$ Int. Work. C. A. Discrete Complex Sys., AUTOMATA 2014.*

[56] Raptis, T. E., "Encoding discrete quantum algebras in a hierarchy of binary words", *$11^{th}$ Vigier Conf.*, Liege, Belgium.

[57] Murthy, C. S. R., Gurusamy, M. (2002) "*WDM Optical Networks, Concepts, Design, and Algorithms*", India:Prentice Hall.

[58] Weinstein, S. B. (2009) "The history of orthogonal frequency division multiplexing" *IEEE Comm, Mag.* 47(11):26-35.

[59] Wooldridge, M. (2002). *An Introduction to Multi-Agent Systems*. J. Wiley & Sons.

[60] Wanhammar, L. (2009) "*Analog Filers using MATLAB*", Springer.

[61] Dimopoulos, H. G. (2012) "*Analog Electronic Filters: Theory, Design and Synthesis*" ACSP-Analog Circuits and Signal Processing, Ed. Ismail, M., Sawan, M. Netherlands:Springer.

[62] Rojas, R. (1996) "Conditional branching is not necessary for universal computation in von Neumann computers" *J. Univ. Comp. Sci.* 2(11):755-768.

[63] Raptis, T. E. (2017) "'Viral' Turing machines, computations from noise and combinatorial hierarchies" *Chaos, Sol. Fract.* 104:734-740.

[64] Martin, J. C. (1976) "Limits of successive convolutions" *Proc. Am. Math. Soc.* 59:52-54.

[65] Borsellino, A., Poggio, T. (1973) "Convolution and correlation algebras" *Kybernetik* 13(2):113-122.

[66] Bostrom, N. " *Superintelligence: Paths, Dangers, Strategies*. Oxford: Oxford Univ. Press.

**Fig. 1**, Points of strict additivity of a bitwise XOR operator ($(\nu \oplus \mu) - \nu - \mu = 0$) form a fractal matrix which is the basis of the *Langlet Transform*, with a known recursive generator.

**Fig. 2,** (a) The set of all 256 arithmetic codes for the truncated rules of all elementary 1D CA, (b) The remaining 80 rules which satisfy the economy principle as explained in section 3.